# Image Retrieval and Pattern Spotting using Siamese Neural Network

Kelly L. Wiggers*, Alceu S. Britto Jr.*¶, Laurent Heutte†, Alessandro L. Koerich‡ and Luiz S. Oliveira§

*Pontifical Catholic University of Parana (PUCPR), Curitiba (PR), Brazil
Email: {wiggers, alceu}@ppgia.pucpr.br

†Normandie Univ, UNIROUEN, UNIHAVRE, INSA Rouen, LITIS, Rouen, France
Email: Laurent.Heutte@univ-rouen.fr

‡École de Technologie Supérieure (ÉTS), Université du Québec, Montréal (QC), Canada
Email: alessandro.koerich@etsmtl.ca

§Federal University of Paraná, Curitiba (PR), Brazil
Email: lesoliveira@inf.ufpr.br

¶State University of Ponta Grossa (UEPG), Ponta Grossa (PR), Brazil

***Abstract*—This paper presents a novel approach for image retrieval and pattern spotting in document image collections. The manual feature engineering is avoided by learning a similarity-based representation using a Siamese Neural Network trained on a previously prepared subset of image pairs from the ImageNet dataset. The learned representation is used to provide the similarity-based feature maps used to find relevant image candidates in the data collection given an image query. A robust experimental protocol based on the public Tobacco800 document image collection shows that the proposed method compares favorably against state-of-the-art document image retrieval methods, reaching 0.94 and 0.83 of mean average precision (mAP) for retrieval and pattern spotting (IoU=0.7), respectively. Besides, we have evaluated the proposed method considering feature maps of different sizes, showing the impact of reducing the number of features in the retrieval performance and time-consuming.**

***Index Terms*—Siamese network, image retrieval, pattern spotting**

## I. Introduction

The large volume of digital information stored in the last two decades by modern society has been the primary motivation for researchers in the Pattern Recognition area to investigate new methods for automatic indexing and retrieval of digital material (images, videos, and audios) based on their contents. Some significant progress may be observed in the field of Content-Based Image Retrieval (CBIR), in which the relevant image candidates must be found in large digital collections based on a given query represented by a whole image or just a pattern available on it [1]–[5]. A more recent and exciting challenge in the CBIR field has been to perform the retrieval process without any contextual information, i.e. without previous knowledge about the patterns to be detected. The lack of information makes unfeasible the use of strategies where specific models are trained based on the patterns (objects) to be retrieved. Such a challenge is common in CBIR solutions devoted to vast libraries of document images, where usually the main tasks are image retrieval and pattern spotting. In the former, the objective is to find every document image that contains the given query, while in the later an additional difficulty is to provide the query location in the retrieved images.

The lack of contextual information mentioned above, combined with the wide variability in terms of scale, color and texture of the patterns present in document collections, which may include seals, logos, faces, and initial letters, makes the definition of a robust representation scheme (feature extraction) a real challenging problem. In the literature, one may find some promising results of traditional representation methods based on local descriptors such as bag-of-visual-words (BoW) [6]. However, some significant contributions have been recently achieved through the use of deep learning based methods, mainly by doing feature extraction using Convolutional Neural Networks (CNN) [7] [8] [9]. Recently, Luo et al. [10] and Wiggers et al. [11] proposed the use of pre-trained CNNs to provide the image representation (feature extraction) achieving very promising results. Different from them, in the present work two deep models (CNN based) are organized in a Siamese architecture. Such a deep architecture has shown successful results in different applications such as face verification [12] [13] and gesture recognition to predict if an input pair of images are similar or not.

In this paper, we address both image retrieval and pattern spotting tasks by using the feature map of a Siamese Neural Network (SNN) trained on the ImageNet dataset to learn how to represent the similarity between two images. With the support of two important concepts, representation and transfer learning, we use the feature map of the trained SNN in our solution to provide the similarity between a given image query and each object candidate (subimage) detected on the document images. The main contributions of our work are threefold: a) evaluation of the SNN similarity representation (feature maps) to perform the retrieval and spotting tasks considering the difficulties inherent to document images; b) evaluation of the transfer learning scheme in which the model trained on regular images of the ImageNet is used without any tuning to perform the retrieval and spotting tasks in the context of document images; c) evaluation of the impact of the SNN feature map size in terms of time-consuming and performance in both tasks, image retrieval and pattern spotting.

A robust experimental protocol using the public Tobacco800 document image collection shows that the proposed method compares favorably against state-of-the-art document image retrieval methods, reaching 0.94 and 0.83 of mean average precision (mAP) for retrieval and pattern spotting, respectively, when considering the retrieval of the Top 5 relevant candidates and IoU=0.7. In summary, the proposed approach succeeds in retrieving relevant image candidates and localizing precisely the position of the searched patterns on it. Additional experiments show that the SNN representation generalizes well, improving the matching performance when compared to deep features of a CNN trained on an image classification task.

The remainder of this paper is organized as follows: Section 2 presents the state-of-the-art techniques in image retrieval and Siamese architectures. The method developed is introduced in Section 3: the representation learning based on SNN, and each step of the image retrieval and pattern spotting tasks. Section 4 presents the Tobacco800 database used for the benchmark, followed by the experimental protocol and results. Finally, Section 5 presents our conclusions and future work.

## II. Related Work

Different techniques can be used to retrieve information from a collection of documents, but they are usually organized in a similar two-step process [14], as follows: a) in an offline step, image candidates are extracted from the document images and indexed using a suitable representation schema (a feature vector); b) in an online phase, given an input query image, a measure of similarity is used to compare it with image candidates extracted from the stored documents, returning a ranked list.

As mentioned before, the main challenges are the lack of previous knowledge about the images to be retrieved and the need for a representation invariant to changes in terms of scale, color, texture, and localization of the query in the image collection. In addition, the query can be presented as a whole document image or a small sub-image (a pattern to be detected). Recently, many papers have been dedicated to image retrieval [15]–[17], while Alaei et al. [18] presented an interesting survey of systems for recovering logos and seals on administrative documents.

Going further in a CBIR system, we can identify three common steps, as follows: a) pre-processing, b) representation (feature extraction), and c) retrieval (matching). In the pre-processing basically, the data collection is prepared for indexing, which is a crucial step for the success of the image retrieval process. An object detection scheme usually based on a sliding window mechanism, or alternative techniques to divide the document image into small regions, results in a set of indexed candidates without the need of knowing their size [19]. One may find in the literature alternative approaches to generate the candidate regions, such as Selective Search [20], Edge Boxes [21] or Bing [22]. An interesting comparison of several algorithms to generate image candidates is presented in Zitnick and Dollár [21]. In their experiments, the Edge Boxes [21] and Selective Search [20] techniques have shown the most promising results in terms of relevant object detection, while Bing has shown to be a faster algorithm but with lower precision.

In the second step (representation), the detected candidates are represented through the use of an appropriate feature extraction method. Here, as in most of pattern recognition fields, deep features based on CNN are an interesting new alternative. In the last few years, deep architectures have been used in image retrieval as feature learning methods [23] [24]. Recently, Gordo et al. [9] evaluated the use of CNN as a feature extractor in several public datasets such as ImageNet, CIFAR-10, and MNIST.

Babenko and Lempitsky [7] explored the use of pre-trained models (transfer learning), observing improvement in retrieval performance. In [25], the authors suggest that feature vectors can be obtained by training Siamese neural networks using pairs of images. The Siamese model has been successfully used to face recognition [26] [13], and signatures or symbol identification [12]. In an SNN, after the convolutional layer, a distance measure between vectors is computed. The name "Siamese" historically comes from the necessity to collect the state of a single network for two different activation samples during training. It can be seen as using two identical, parallel neural networks sharing the same set of weights [27]. The first idea of Siamese network was published by Bromley et al. [28] for signature verification problems.

Qi et al. [29] also proposed a sketch-based image retrieval to pull output feature vectors closer for input pairs that are labeled as similar and push them away if irrelevant. Although this approach is related to our work, the proposed architecture and image search are different. In our method, we decide to evaluate the generalization of a model trained on ImageNet, avoiding a fine-tuning on the used document images.

A recent work proposed by Chung et al. [30] focused on a deep Siamese CNN pre-trained on ImageNet dataset. The proposed method computed contrastive loss function and showed good performance in Diabetic Retinopathy Fundus image dataset. However, the authors use only binary image pair information. In contrast, in document images, the features can be more complex, with variation in shape, color, and texture. Lin et al. [31] propose a Siamese model to learn a feature representation and finding matches between street view and aerial view imagery. However, they use their dataset to learn the representation which is not feasible in our case.

The third step of a CBIR system (retrieval) usually requires a distance metric to compute the similarity between the feature vectors representing the query image and the list of image candidates. It is common to use a distance like Euclidean or Cosine, or to use a trained model (as an SNN for example) to compute the similarity between two images.

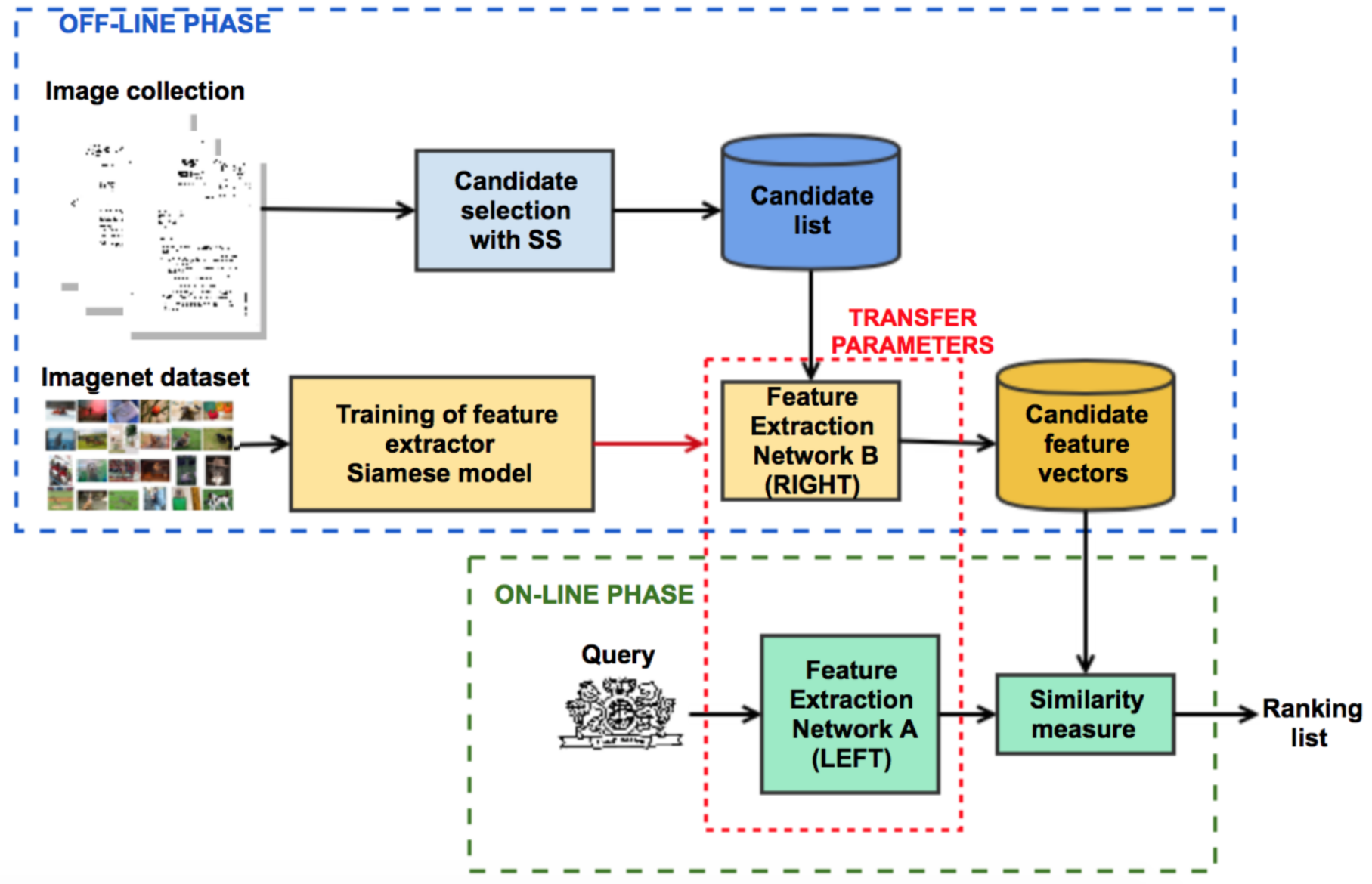

Fig. 1: Overview of the proposed image retrieval method

## III. Proposed Method

Motivated by the significant growth of digital content storage observed in the last years, we propose here a new method that contemplates both image retrieval and pattern spotting, which are the main tasks of an image search engine. The image retrieval consists of finding every document that contains a given query, while additional information is necessary for the pattern spotting, which is related to the exact location of the query in the retrieved images. Figure 1 presents a general overview of the proposed SNN based method. In the offline phase, the Selective Search (SS) algorithm detects the possible pattern locations, generating sub-images that are candidates to be compared with the queried images. Still, in the offline phase, the SNN is trained on a subset of images from the ImageNet dataset organized as image pairs (similar and non-similar). The idea is to learn how to represent the similarity between two images. With this strategy, we can avoid any additional training usually performed to fine-tuning the model on a specific document data collection. Using the concept of transfer learning, the SNN is used as a feature extractor. This is done for saving time, the feature map of each image candidate is stored to be further used in the online phase. In the search process carried out during the online phase, given an image query, an exhaustive scheme is performed considering the whole list of generated candidates. The final output is a ranked list of relevant candidates. Each step of the proposed method is detailed in the next sections.

**Pre-processing: generating object candidates**: the SS algorithm [20] is used to execute an exhaustive search segmenting the document image into regions of interest (image candidates). In this paper, we modify the parameters for object location, considering an adaptive threshold scheme to change the block size and offset (constant subtracted from the neighborhood mean) values. Thus, the parameters were defined as follows: *block*=241 and *offset*=0.12, *k*=50 and 100, *color+texture+filler+size*. With this configuration, we observed an increase in the quality of the candidates generated.

**Siamese network**: inspired by Melekhov et al. [32], and Chung and Weng [30], we used the network illustrated in Fig. 2a to learn a deep representation for distinguishing similar and non-similar image pairs. The network is based on the AlexNet architecture, proposed by Krizhevsky et al. [33] which is composed of 8 layers, 5 convolutional layers followed by 3 fully connected layers. The AlexNet has been considered as a good starting point for the proposed method. It is easy to implement and has shown to be effective in different deep learning scenarios, showing a good compromise between the number of layers and final accuracy. Based on our previous work [11], we have considered an additional layer in the AlexNet architecture to evaluate feature maps of different sizes.

As we can see in Fig 2a, the proposed SNN architecture is composed of two identical Alexnet trained on the ImageNet dataset. The last fully connected layer with the number of categories was removed and the fully connected layer with $n$ dimensions was trained for feature representation. Different from the conventional SNN, here the Euclidean distance is implemented with the proposed layers available in the Caffe tool [34], as follows: *Eltwise* to compute the distance between the feature map entries; *Power* to compute the square root; and *Reduction* to sum the obtained distances reducing it to a scalar. After that, we implement a fully connected layer with one output and loss layer. As shown in Fig. 2a, our SNN architecture has the distance layers isolated from the loss layer. It may facilitate the use of the distance value in the online phase of the method. We set the learning rate in $10^{-3}$ with an exponential function. The activation function used was Stochastic Gradient Descent (SGD), as recommended by Recht et al. [35].

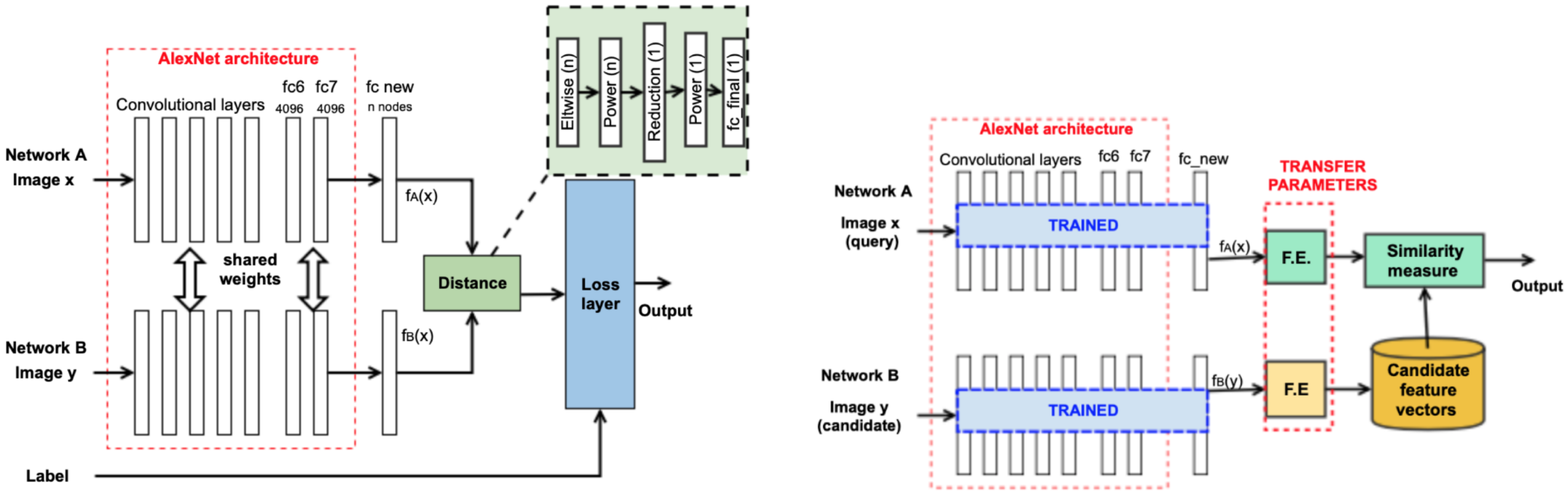

(a) Training (b) Testing

Fig. 2: Overview of our Siamese network for document image retrieval

During the training, the input is a pair of images $x$ and $y$, where $X$ and $Y$ are ImageNet positive and negative pairs in the training set, respectively. The inputs $x$ and $y$ are fed into the two Alexnet networks $A$ and $B$, both with an additional layer with $n$ dimensions. The images of ImageNet used to train the SNN were 100,000 similar pairs and 150,000 non-similar pairs. We generated 1.5 $\times$ more non-similar images for training and test as recommended by Melekhov et al. [32]. We split the images into training (70%) and test (30%) subsets. The output distance of $A$ and $B$ is fed to a *sigmoid cross entropy loss* layer, that aims to minimize the difference of the probability distribution between the predicted labels and ground truth labels [36]. For each sample we have:

$$L_i = \sum_{k=1}^{C} t_k(y_i) \log P(y_i = k | b_i; w_k) \tag{1}$$

where $t_k(y_i)$ is the distribution of ground truth labels $y_i$; $P(y_i = k|b_i; w_k)$ the probability distribution of predicted labels. The sigmoid function is used for the two-class logistic regression, that is, when using a network, we try to get 0 or 1.

During the testing step (see Fig. 2b), we use the concept of transfer learning to apply the previously trained SNN network as a feature extractor to retrieve document images. As one may see, for the sake of time, we use the proposed model as a feature extractor, using an external distance measure during the online phase as shown in Fig. 2b. Thus, each image candidate is processed only once.

**Image retrieval task**: an exhaustive search based on Euclidean distance is used providing a rank with the Top-$k$ candidates for each query (see Fig. 3), as already used in [11]. Average Precision (AP) and Recall for each query were adopted as performance measures.

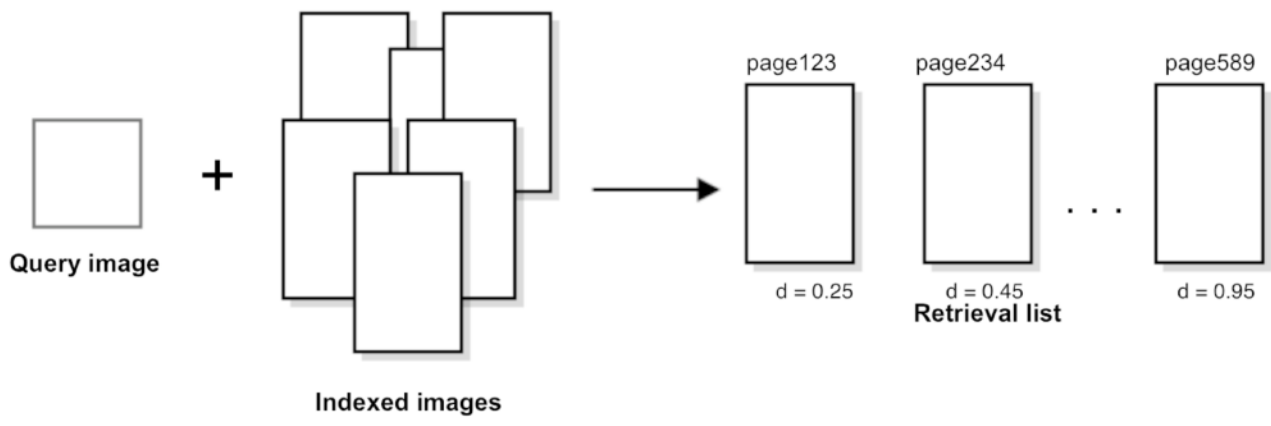

Fig. 3: Example of the results shown in the image retrieval task. The return is a list of non-repeated images, ordered by distance. [11]

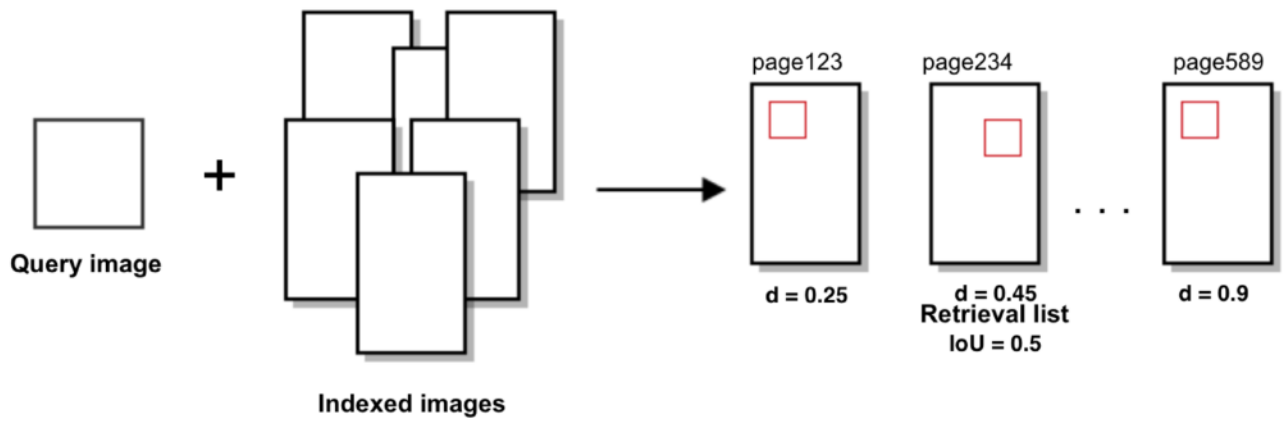

Fig. 4: Example of the results presented in the pattern spotting task. The output is a list of images, ordered by distance and their location.

**Pattern spotting task**: here the pattern (query) location(s) must be provided on each image document. The overlap between the query and the candidate (region of interest) retrieved is computed, as shown in Fig. 4 (red square). The overlap estimation considers the position of the query ($x$,$y$), its corresponding area $q_1$, the positions of the candidate ($x_1$,$y_1$), and its corresponding area $o_1$, as denoted in Equation 2 [37].

$$\text{IoU}(x, y) = \frac{q_1 \cap o_1}{q_1 \cup o_1} \qquad (2)$$

The relevance of a candidate is related to its overlap with the query. For this purpose, the Intersection over Union (IoU) is computed. IoU=0.7 is considered to be a reasonably good result, since IoU=0.5 is considered too small and IoU=0.9, very restrictive [21]. However, we consider the analysis: 0.1≤IoU≤0.7 in order to determine that a positive candidate is retrieved, and at the end, the precision and the recall are calculated. Finally, the mAP is calculated to evaluate the results considering all the queries.

## IV. Experiments

This section describes our experiments on the Tobacco800 dataset. This public subset of the Complex Document Image Processing (CDIP) collection was created by the Illinois Institute of Technology [38]. The 1,290 document images were labeled by the Language and Media Processing Laboratory (University of Maryland). It contains 412 document images with logos and 878 without logos. In our experiments, we have considered the 21 categories presenting two or more occurrences, making 418 queries for the search process.

### A. Experimental Protocol

In the pre-processing stage, the improved Selective Search algorithm produces 1.2M of regions of interest considering the 1,290 documents in the database. All these images were considered in our experiments. The improved version of SS increased in 13× the quality of objects that overlap the query in more than 90%, when compared to our previous work [11]. In fact, we have considered the **aspect ratio** of the query image to guide the SS algorithm. The candidate $c$ will only be considered for retrieval if the aspect ratio $\frac{height_c}{width_c}$ has a maximum of 25% difference with the query $q$. For example, if a query has $\frac{height_q}{width_q}$ =0.5, the candidate similarity calculation will only be performed if the candidate has a $\frac{height_c}{width_c}$ between 0.375 and 0.625. For Tobacco800 dataset, by applying this query based contextual information, the initial 1.2M of candidates were reduced to 873,876 candidates, improving the quality of the pre-processing stage (candidate generation).

The SNN architecture trained with ImageNet dataset provides 4096-dimensional feature maps corresponding to its original $fc_7$ layer. We have also investigated the addition of a new layer ($fc_{new}$) to reduce the feature maps to 512, 256 and 128 dimensions. These values were used in [11] to reduce the dimensionality of the feature map of a CNN model. In addition, we decided to use the SNN as a feature extractor to avoid spending time to process each candidate image more than once. Table I compares the time spent in the retrieval task for some queries when using the distance layer implemented inside the SNN, and the SNN used as a feature extractor. In the last, the feature map is extracted using the fully-connected layer, then an external Euclidean distance measure is computed. We observed that the computational time to retrieve a single query using the SNN as a feature extractor is about 11 times lower.

TABLE I: Computational time for the retrieval task (in seconds). Distance layer of Siamese model .vs. Siamese model as a feature extractor. Model with 256-feature map

| Query Image | # of Candidates | Siamese Distance Layer | Siamese as Feature Extractor |
|---|---|---|---|
| | 210,148 | 571.43 | 52.08 |
| | 124,060 | 427.56 | 35.29 |
| | 167,571 | 492.16 | 54.02 |

The performance of both tasks, image retrieval and pattern spotting, is evaluated using the mean Average Precision (mAP) and Recall for all queries. In the experiments, we evaluated the effect of the choices made for the three mentioned method stages, and we compared the final results with our previous method [11] and the current state-of-the-art.

TABLE II: mAP with Euclidean distance, Top-$k$ ranking, and Siamese with feature map sizes $n = \{4096; 512; 256; 128\}$

| Feature Map | Top-*k* | | | | |
|---|---|---|---|---|---|
| | 5 | 10 | 25 | 50 | 100 |
| 4096 | **0.944** | **0.872** | **0.735** | **0.546** | **0.306** |
| 512 | 0.744 | 0.640 | 0.486 | 0.358 | 0.220 |
| 256 | 0.746 | 0.654 | 0.503 | 0.368 | 0.227 |
| 128 | 0.738 | 0.655 | 0.506 | 0.373 | 0.229 |

### B. Results and Discussion

Table II shows the experimental results of the proposed image retrieval method, comparing the SNN with a 4096-dimensional feature map with those with $n$ entries, where $n$=$\{512, 256, 128\}$. The similarity between the network feature maps was calculated using the Euclidean distance. Afterwards, the Top-$k$ candidates were chosen to generate a ranked list of relevant image candidates according to the mAP, where $k$=$\{5, 10, 25, 50, 100\}$. As one may see in Table II, the best results were achieved by using the AlexNet with 4096 features resulting in 0.94 and 0.87 of mAP in the Top-5 and Top-10 ranking, respectively. The recall is higher than 90% in Top-5 and Top-10. With the additional layer, the best results are related to 256 features with a 0.74 of mAP for Top-5 ranking. The performance is significantly better using the 4096-dimensional feature map than any version of the additional layer. However, the computational cost to index all the vectors of the dataset is much higher.

Table III shows some qualitative results of the logos retrieved using the SNN with full feature map (4096 dimensional). These results are very promising since many correct logos were retrieved in the Top-5 ranking. We can observe the good performance especially in the fifth row, where the query is very similar to a signature, but we did not have false positives in Top-5 ranking. In the last row we can see some false candidates, motivated by the presence of very few positive samples in the dataset for this query. We can also see the results per category in Fig. 5. We observed in some categories an mAP greater than 0.8. The categories 8, 9, and 16, in particular, have many samples in the dataset. The categories 5, 17, and 19 have very few samples that are moreover, of bad quality. In our previous approach [11], several queries had 0.0 of mAP in some categories, while in this new approach we do not observe queries with a mAP lower than 0.2.

TABLE III: Qualitative retrieval results. Query and the first five retrieved logos using 4096-dimensional feature map

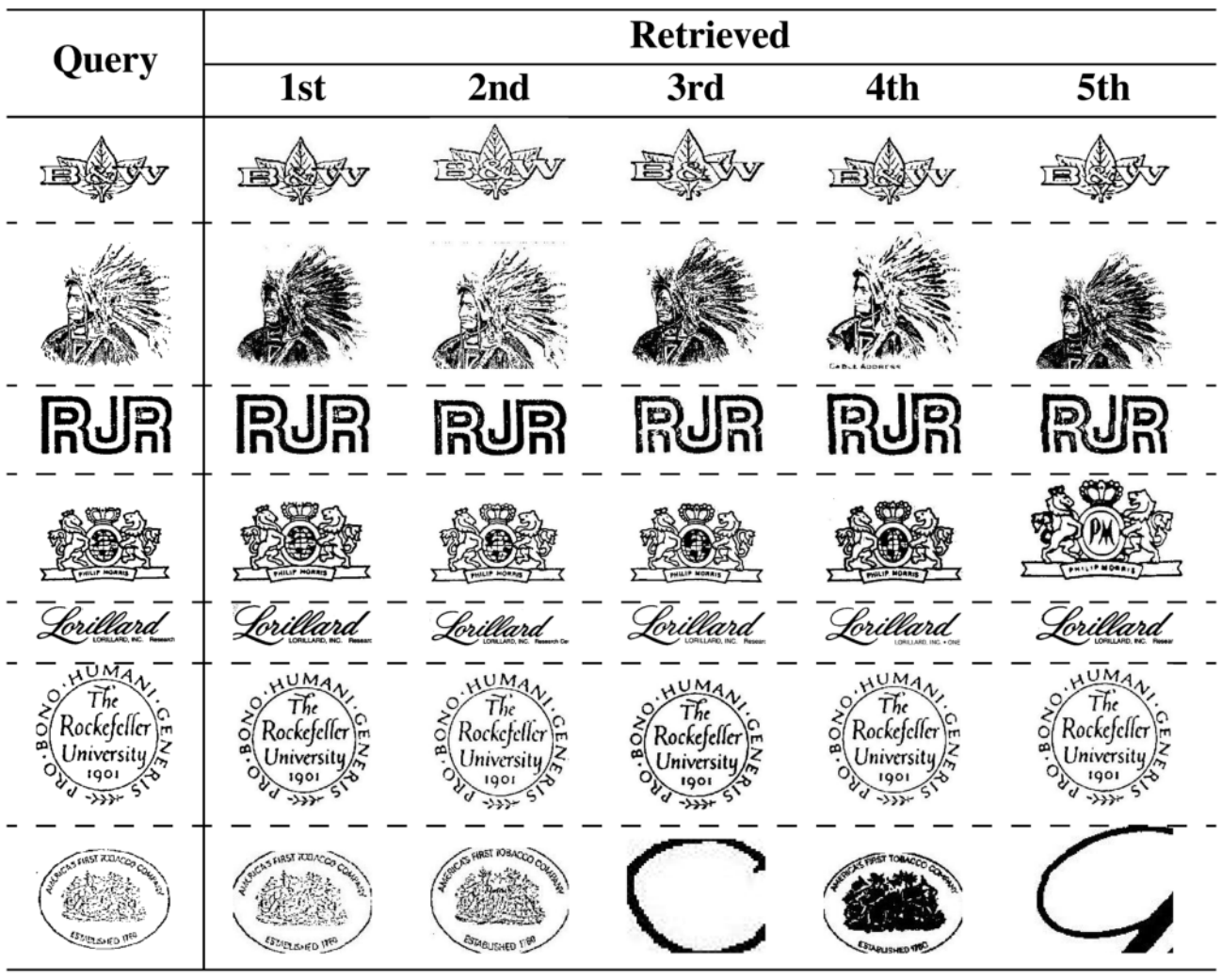

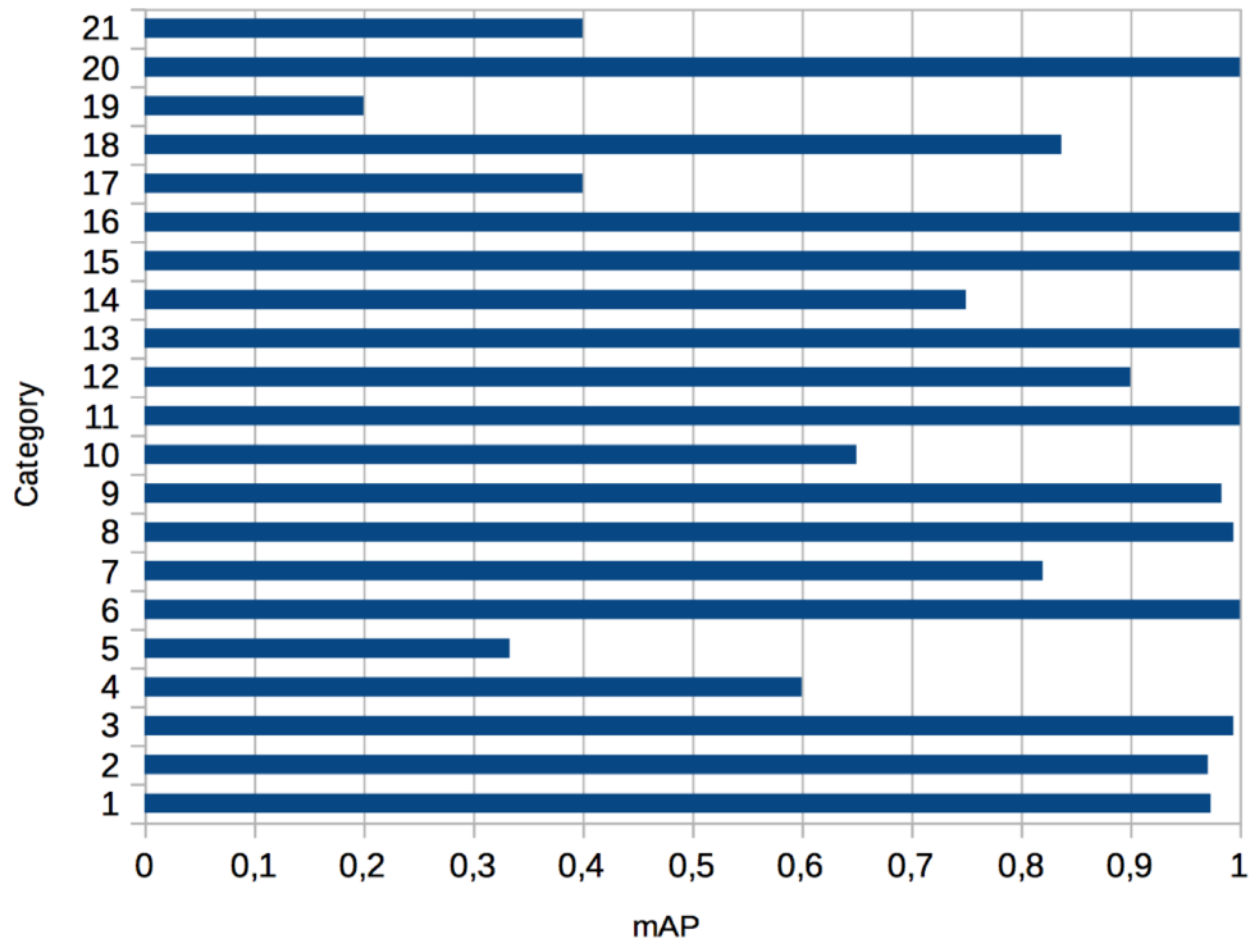

Fig. 5: Results by category using a Siamese network with 4096-dimensional feature map and Top 5 ranking

The experiments were compared with the state-of-the-art and the results are shown in Table IV. If the results of Jain and Doermann [39] are considered, the method proposed in this paper is 93.8%, 42.22%, 45.33% and 45.50% better using the SNN with 4096, 512, 256 and 128 feature map, respectively. The proposed method is 2.08% better considering the SNN with 4096 entries when compared to the results achieved by Rusinol and Llados [40]. However, the results are 32.50%, 30.13%, 24.80% below for 512, 256 and 128 feature map, respectively. Comparing the results with our previous work in Table V, for the 4096-dimensional and 512-dimensional feature maps the proposed SNN-based method is 28.23% and 20.75% better for Top-10, respectively. However, considering

TABLE IV: Comparison with the state-of-the-art retrieval task using Euclidean distance and Top-10 and Top-25

| **Author** | **Top-*k*** | |
|---|---|---|
| | **10** | **25** |
| Jain and Doermann [39] | 0.450 | NA |
| Rusinol and Lladós [40] | NA | 0.720 |
| 4096-dimensional feature map | **0.872** | **0.735** |
| 512-dimensional feature map | 0.640 | 0.486 |
| 256-dimensional feature map | 0.654 | 0.503 |
| 128-dimensional feature map | 0.655 | 0.506 |

NA: Not Available

TABLE V: Comparison with the state-of-the-art retrieval task using Euclidean distance and Top-*k* ranking for 4096, 512, 256 or 128-dimensional feature map.

| **Author** | **Feature Map Dimension** | **Top-*k*** | | | |
|---|---|---|---|---|---|
| | | **10** | **25** | **50** | **100** |
| Wiggers et al. [11] | 4096 | 0.68 | 0.57 | 0.45 | 0.32 |
| | 512 | 0.53 | 0.40 | 0.31 | 0.23 |
| | 256 | **0.72** | **0.61** | 0.50 | **0.35** |
| | 128 | 0.69 | 0.61 | **0.51** | 0.34 |
| Proposed Method | 4096 | **0.87** | **0.73** | **0.54** | **0.30** |
| | 512 | 0.64 | 0.48 | 0.35 | 0.22 |
| | 256 | 0.65 | 0.50 | 0.36 | 0.22 |
| | 128 | 0.65 | 0.50 | 0.37 | 0.22 |

TABLE VI: Computational time for the retrieval task (in number of candidates processed per second)

| **Siamese Model** (feature map dimension) | **Feature Extraction** (cand/sec) | **Retrieval** (cand/sec) |
|---|---|---|
| 4096 | 70.38 | 65.02 |
| 512 | 76.87 | 233.50 |
| 256 | 76.26 | 392.22 |
| 128 | 76.34 | 451.55 |

TABLE VII: Pattern spotting results with Euclidean distance considering different values of IoU and Top-5

| **Siamese Model** | **IoU** | | | | |
|---|---|---|---|---|---|
| (feature map dimension) | **0.1** | **0.3** | **0.5** | **0.7** | **0.9** |
| 4096 | **0.841** | **0.841** | **0.839** | **0.836** | **0.793** |
| 512 | 0.711 | 0.707 | 0.705 | 0.699 | 0.607 |
| 256 | 0.719 | 0.716 | 0.714 | 0.707 | 0.614 |
| 128 | 0.706 | 0.704 | 0.702 | 0.696 | 0.616 |

TABLE VIII: Qualitative spotting results for some queries. Query logo and first retrieved logo by similarity using 4096-dimensional feature map and location (red square)

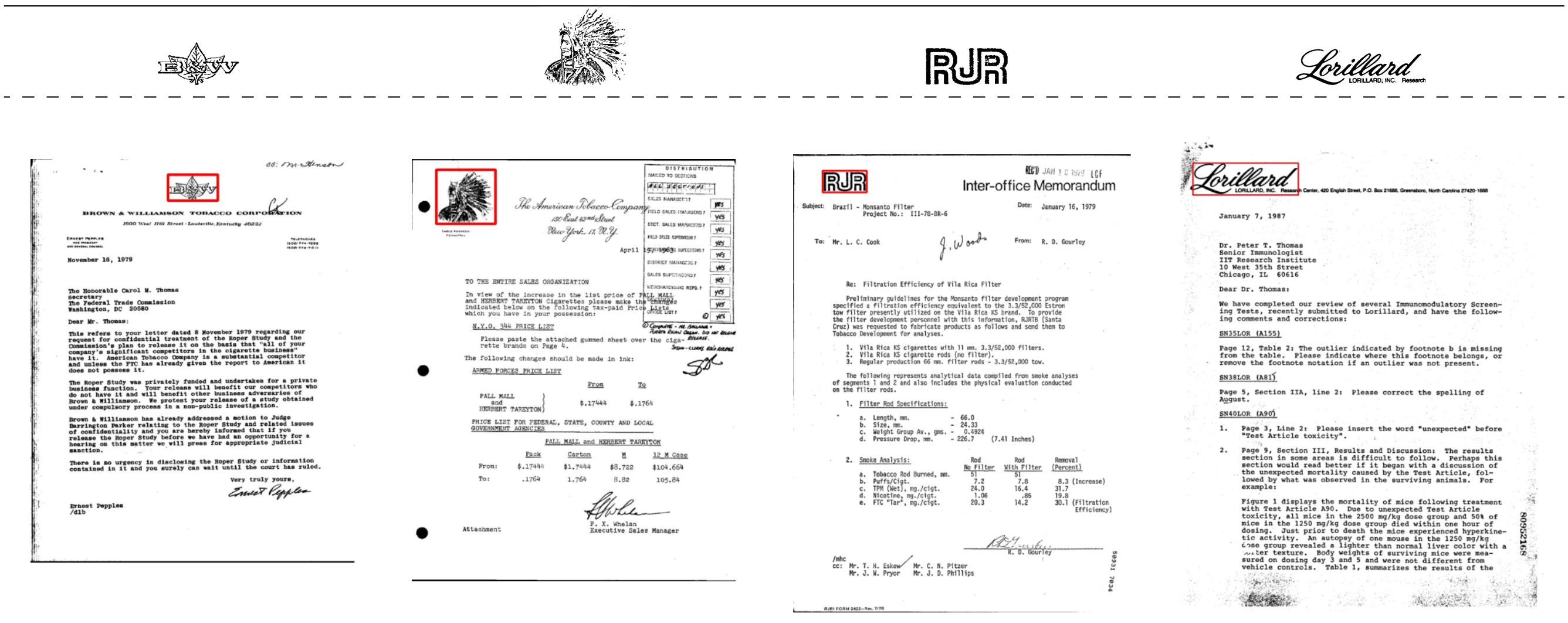

TABLE IX: Comparison with the state-of-the-art for Pattern Spotting. Top-5, IoU≥0.6, classes with at least three samples

| Approach | mAP | Recall (%) |
|---|---|---|
| Le et al. [41] | **0.970** | 88.42 |
| Le et al. [42] | 0.910 | 88.78 |
| Siamese Model (4096) | 0.922 | **92.72** |
| Siamese Model (256) | 0.718 | 73.19 |

the 256-dimensional and 128-dimensional feature maps, the results were 9.20% and 5.07% worst, respectively.

Finally, the computational time of the proposed method was evaluated. The computational resources consist of 1 AMD Ryzen 5 1600, 32GB of RAM, 1 GPU NVIDIA GTX 1080 with 8GB of memory and 2560 CUDA. Table VI compares the number of candidates processed per second by each approach (512, 256 and 128 dimensions) and the original feature map (4096 dimensions). As one may see, the number of candidates processed per second increased by approximately 6.10%, 9.20% and 8.40% when considering the feature maps with 512, 256 and 128 dimensions, while in the retrieval process it increased in almost 2, 3.5 and 5 times, respectively.

Table VII shows the experimental results of the pattern spotting task. Here, the candidate is taken as relevant if it overlaps enough with the image query. We can observe that our best results were achieved using the whole feature map (4096), but the results related to the reduced maps are quite competitive. Despite the reduction of about 0.12 in mAP, they have shown a significant reduction in terms of time consuming as shown in Table VI. Still, in Table VII, we can see a small gap in terms of localization performance between IoU=0.1 and IoU=0.7. This means that our system succeeded not only to retrieve the relevant image candidates for each query, but also in finding the query position precisely. A qualitative analysis can be seen in Table VIII, in which we have some queries and the first candidate retrieved with its respective location.

Table IX shows a comparison with the current state-of-the-art for pattern spotting on the Tobbaco dataset. For such a comparison we have used the same experimental parameters of [41], [42], which considers Top-5, IoU≥0.6 and classes with at least three samples per category. The mAP achieved by the proposed Siamese model with 4096 is 1.31% better than mAP achieved by Le et al. [42] but it did not outperform the mAP presented in [41]. On the other hand, the recall is 4.44% and 4.86% better than [42] and [41] respectively. However, it is important to highlight that both [41] and [42] require the previous knowledge of the logo gallery, what is not necessary for the proposed method.

## V. Conclusion

This paper presented a novel approach for image retrieval and spotting in document images where the images are represented using a Siamese model trained on the ImageNet dataset. We evaluated the results of conventional AlexNet architecture and a modifier version with an additional layer with the aim of reducing the dimension of the feature map. Our experimental results were very promising. It was possible to observe an increase in the mAP, since the features generalize well and improve the matching performance compared to features obtained with networks trained for image classification.

Further work must be carried out to improve the SNN model by using other architectures to avoid exhaustive comparisons of feature maps, allowing the use of the model distance layer during the online phase. Besides, we plan to evaluate the proposed SNN model in different datasets, such as in Paris, INRIA Holidays and DocExplore.

## Acknowledgment

CAPES (Coord. for the Improvement of Higher Education Personnel) and CNPq (National Council for Scientific and Technological Development) grant 306684/2018-2.